\documentclass[letterpaper, 10 pt, conference]{ieeeconf}  
\IEEEoverridecommandlockouts                              
\usepackage{subfigure}
\usepackage{booktabs}
\usepackage{balance}
\usepackage{xcolor}
\usepackage{tabularx}
\usepackage{graphicx,dblfloatfix} 
\usepackage{amsmath}
\usepackage{amssymb}
\usepackage{listings}
\usepackage{textcomp}
\usepackage{url}
\usepackage{multirow}
\usepackage{todonotes}
\usepackage{hyperref}
\usepackage{cleveref}
\usepackage{wrapfig}
\crefformat{footnote}{#2\footnotemark[#1]#3}
\usepackage{ragged2e} 
\usepackage[subnum]{cases}
\newcolumntype{Y}{>{\RaggedRight\arraybackslash}X} 
\usepackage{algorithm}
\usepackage[noend]{algpseudocode}

\usepackage{tikz}
\usepackage{tkz-euclide}
\usetikzlibrary{math}
\usepackage{pgfplots}
\usetikzlibrary{arrows}

\usepackage{float}

\usepackage{multicol, blindtext}

\usepackage{multirow}
\usepackage{adjustbox}
\usepackage{booktabs}
\usepackage{footnote}

\newcommand{\alignedequation}[2]{ %
    \begin{equation} \label{#1} %
    \mbox{\fontsize{9}{9}\selectfont\(\begin{aligned}#2\end{aligned}\)} %
    \end{equation} %
} 

\usepackage{bm}

\title{\LARGE \bf
   Semi-supervised Gated Recurrent Neural Networks for Robotic Terrain Classification  
}

\author{Ahmadreza Ahmadi$^{\dagger}$, T{\o}nnes Nygaard$^{\ddagger}$, Navinda Kottege$^{\dagger}$, David Howard$^{\dagger}$, Nicolas Hudson$^{\dagger}$  
\thanks{$^{\dagger}$ A. Ahmadi, N. Kottege, D. Howard, and N. Hudson are with the Robotics and Autonomous Systems
Group, CSIRO, Pullenvale, QLD 4069, Australia.All correspondence should be addressed to {\tt\small ahmadreza.ahmadi@data61.csiro.au}}%
\thanks{$^{\ddagger}$T. Nygaard is with the University of Oslo, Norway.}
}
\begin{document}
\maketitle
\thispagestyle{empty}
\pagestyle{empty}

\begin{abstract}
Legged robots are popular candidates for missions in challenging terrains due to the wide variety of locomotion strategies they can employ. Terrain classification is a key enabling technology for autonomous legged robots, as it allows the robot to harness their innate flexibility to adapt their behaviour to the demands of their operating environment. In this paper, we show how highly capable machine learning techniques, namely gated recurrent neural networks, allow our target legged robot to correctly classify the terrain it traverses in both supervised and semi-supervised fashions. Tests on a benchmark data set shows that our time-domain classifiers are well capable of dealing with raw and variable-length data with small amount of labels and perform to a level far exceeding the frequency-domain classifiers. The classification results on our own extended data set opens up a range of high-performance behaviours that are specific to those environments. Furthermore, we show how raw unlabelled data is used to improve significantly the classification results in a semi-supervised model.  

\end{abstract}

\section{Introduction}
\label{sec:introduction}

Bio-inspired legged robots offer advantages when walking in extreme environments with their ability to adapt to instantaneous conditions including undulation, slope, roughness, and terrain types. This is possible by changing gaits, foot-tip arc shapes, footfall placement, stride length, etc., to  tune their behaviour and overcome the challenges presented by their environment.  Compared to other types of robots, they have more flexibility to effectively couple their hardware and software configuration to the specifics of the terrain.  

An important step to fully harnessing these myriad degrees of behavioural freedom is {\em terrain classification}; the ability for a robot to correctly gauge the type of terrain it is on, and thus enact an appropriate response to overcome the challenges of that terrain.  A plethora of previous approaches focus on terrain classification with legged robots using various methods with varying levels of accuracy~\cite{hoepflinger2010haptic, filitchkin2012feature, best2013terrain, christie2016acoustics, wu2016integrated}.

\begin{figure}[t]
  \centering
  \includegraphics{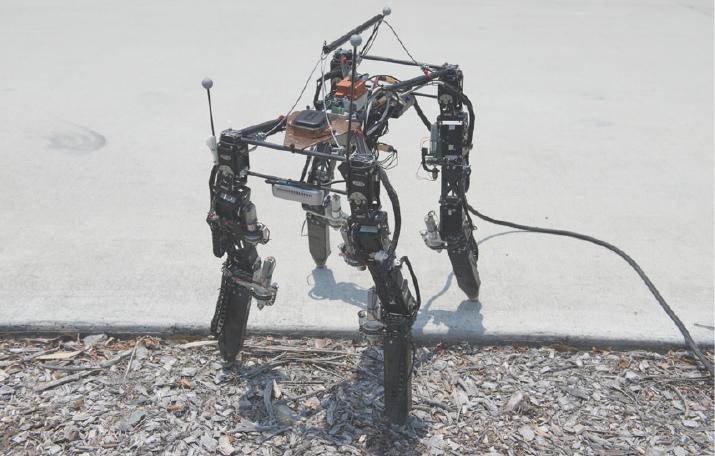} 
  \caption{DyRET quadruped robot on outdoor terrain.}
  \label{fig.dyret}
\end{figure}

In this work, we focus on the use of modern, highly capable deep learning methods, namely {\em gated recurrent neural networks}, to perform this classification in both supervised and semi-supervised schemes. Our main contributions are:

\begin{itemize}
       \item A deep learning model for terrain classification via proprioceptive sensing, which significantly outperforms state of the art frequency-domain approaches.
       \item Comparison of two gated RNN models (LSTMs and GRUs) for terrain classification 
       \item Extensive testing, both on the previously-introduced benchmark PUT dataset~\cite{bednarek2019touching, bednarek2019robotic} which covers indoor terrains, and another large, varied outdoor dataset collected by the authors.
       \item The first semi-supervised model for robotic terrain classification capable of dealing with raw and variable-length data. We show comparable performance to fully supervised methods, whilst requiring much less annotated data. 
\end{itemize}

This work covers three entries of a recent collection of Grand Challenges~\cite{Yangeaar7650} \footnote{Biohybrid and bioinspired  robots, Navigation and exploration in extreme environments, and Fundamental aspects of artificial intelligence (AI) for robotics}. Our approach significantly outperforms the frequency-domain models in legged terrain classification~\cite{bednarek2019touching}, and offers the potential for legged robots to generate high-performance behavioural responses that are customised to their environments. Moreover, to the best of our knowledge, this is the first study that investigates a semi-supervised model that can directly use raw and variable-length time-series data for robotic terrain classification. Our dataset\footnote{\url{https://doi.org/10.25919/5f88b9c730442}} and code\footnote{\url{https://github.com/csiro-robotics/deep-terrain-classification.git}} are available online.

In the remainder of the paper we discuss background research (Section~\ref{sec:related works}), introduce our target robot and data sets (Section~\ref{sec:Robots and Datasets}), and describe our method and results in Sections~\ref{sec:Method} and~\ref{sec:Results} respectively.  We conclude with a discussion of our results, and directions for future research in Section~\ref{sec:conclusions}.

\section{Related work}
\label{sec:related works}

   


Legged robot locomotion in rough and unstructured terrain can be addressed either by purely reactive methods or deliberative methods. Reactive methods \cite{Kottege2015Energetics,bjelonic2016proprioceptive,williamson2016terrain,Homberger2016Terrain-dependant,bellicoso2016perceptionless} adapt to new terrain types by changing locomotion parameters or body configuration accordingly. While these methods typically incur low processing overheads, the actual adaptation process can be slow as the robot has to walk over the new terrain for the changes to trigger. Therefore, there is a risk of the robot entering terrain beyond it's locomotion capability before it is identified, potentially stranding the robot in a local minima. 

Deliberative methods first classify the type of terrain, then switch to appropriate behaviours rather than adapting reactively. The first step in this process is terrain classification \cite{hoepflinger2010haptic, filitchkin2012feature, best2013terrain, christie2016acoustics, wu2016integrated}. Proprioceptive sensing and visual perception are often used to gain information about the terrain through the robots' joints, force sensors attached to the feet, inertial measurement units (IMUs), and cameras.  These input sequences are then used in a machine learning framework to classify the type of terrain the robot is traversing. Methods based on visual perception may fail in the presence of large variations in illumination intensity, dust, fog, smoke, or seasonal changes in color perception, etc.  The state of the art in proprioceptive sensing is dominated by frequency domain classifiers \cite{valada2018deep, bednarek2019touching, bai2019deep}.  Variable length input sequences are encountered especially with legged robots, e.g., when the robot walks at different speeds resulting in variable step frequencies. If this is not addressed, the classification results will only be valid when the robot is walking through terrain at the speed used during the training phase. Recent neural approach \cite{bednarek2019touching} used an FFT-based pre-processor to overcome problems faced by neural network-based methods when processing variable length sequences. 

Recurrent Neural Networks (RNNs) have been widely used in natural language processing tasks as a next step predictor, meaning that they predict what comes next in a sequence~\cite{sutskever2011generating, graves2013generating, zaremba2014recurrent, evermann2017predicting}. Here, RNNs are used in an unsupervised paradigm as data labelling is not required during  training.  Semi-supervised learning combines supervised and unsupervised learning to deal with the major drawback of supervised models, an extensive reliance on hand-annotated datasets. Semi-supervised learning has been widely studied for image classification, natural language processing, and video prediction, and achieved outstanding results with small amounts of labelled data~\cite{kingma2014semi, dai2015semi, rasmus2015semi, maltoni2016semi, lotter2016deep, berthelot2019mixmatch}. 

Applications of semi-supervision for robotic terrain classification are underrepresented in the literature. A semi-supervised Laplacian Support Vector Machine (SVM) was recently proposed for a relatively small dataset (1584 samples for 6 different types of terrains); it was shown that the proposed method could achieve higher accuracy than a traditional Laplacian SVM~\cite{shi2020laplacian}. Ten time-domain features were used as input, such as number of sign variations in a sample, sample mean, sample variance, an auto-correlation function of a sample, an impulse factor,  etc.  In other words, the method does not deal with raw data, and requires many hand crafted features to function. Furthermore, the proposed method cannot deal with variable-length datasets such as PUT and QCAT because the length of data needs to be set in advance to extract time-domain features.  

We are therefore motivated to investigate the utilisation of RNNs as semi-supervised models for robotic terrain classification, to deal with raw and variable-length time-series data.  We propose a semi-supervised method that  incorporates unsupervised RNNs and supervised RNNs, and test on two datasets (one indoor, one outdoor).  We demonstrate (i) our time domain representation outperforms recent frequency-domain representation, and (ii) semi-supervision provides equivalent performance to state of the art algorithms, whilst requiring much less labelled data.

\section{Robots and Datasets}
\label{sec:Robots and Datasets}

In this section we introduce our robotic test platform, together with the PUT~\cite{bednarek2019touching} and QCAT datasets used in the experimentation.  Details of the two datasets are provided in Table~\ref{tab1}.

\subsection{PUT Dataset (indoors)}
The PUT dataset~\cite{bednarek2019touching} was collected using a six-legged robot platform equipped with a Force/Torque sensor on one of its legs, sampled at 200\,Hz. The robot walked at three different speeds, in six walking directions, on six different artificial indoor terrains (sand, rubber, concrete, artificial grass, wood chipping and gravel).  Eighty steps are completed for each combination of speed, direction, and terrain, giving a total of 8640 steps in the whole dataset.  

\subsection{Test Platform}
We used the open-source Dynamic Robot for Embodied Testing (DyRET) platform~\cite{nygaard_ICRA19} (Fig.~\ref{fig.dyret}) to collect the data.  DyRET is a mammal-inspired robot weighing about 5\,kg, built specifically to facilitate machine learning research on real-world robots~\cite{nygaard19exp}.  Control is via position controlled Robotis servomotors and a high level spline-based gait controller~\cite{ny2019evolving}.  The robot is equipped with individual directional force sensors (OptoForce OMD-20-SH-80N) on each foot, reporting force on three axes at 100\,Hz.  An Attitude and Heading Reference System (XSens MTI-30) reports linear acceleration, rotational velocity and orientation at 100\,Hz.

\subsection{QCAT Dataset (outdoors)}
\begin{figure*}
  \centering
  \includegraphics{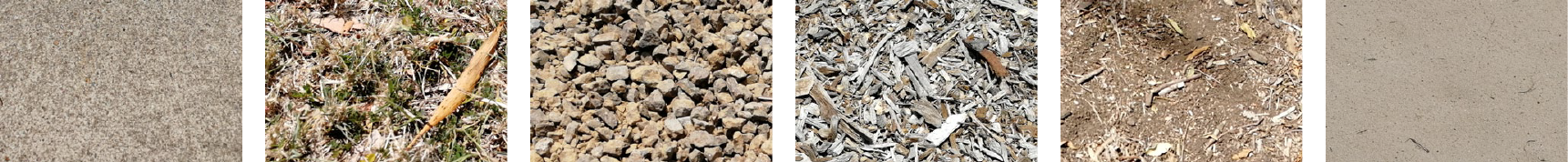} 
  \caption{The different terrains used for collecting our dataset (QCAT). Left to right: Concrete, Grass, Gravel, Mulch, Dirt and Sand.}
  \label{fig.terrains}
\end{figure*}

The QCAT dataset was collected at different locations on CSIRO's QCAT site in Brisbane, Australia, in November 2019.  Fig.~\ref{fig.terrains} shows the different environments comprising the data set: a concrete road, grass, gravel, mulch, a dirt path, and sand.
Data collection involved walking with a fixed gait, but with three different step frequencies (0.125\,Hz, 0.1875\,Hz and 0.25\,Hz) and two different step lengths (80\,mm and 120\,mm), for a total of six different speeds tested per surface.
The robot walks forwards for eight steps, with ten repeats for a total of 80 steps per speed and surface.
To be representative of the terrain type, each repeat occurs on a different part of the terrain. 
The dataset consists of the force senors' measurements  (12 dimensions: 4 sensors $\times3$) and the IMU sensor's measurements (10 dimensions: 3 of linear accelerations, 3 of angular velocities, and  4 of orientations). 



\begin{table}
    \caption{Statistics for the two datasets used in this paper. DyRET is symmetric with sensors on both front and back feet.The data represents both forward and reverse walking directions.}
    \centering
    \begin{tabular}{lrr}
    \toprule
                                  & QCAT dataset & PUT dataset \\
    \midrule
        Number of surfaces    &           6 &           6 \\
        Number of speeds      &           6 &           3 \\
        Number of directions  &          *1 &           6 \\
        Sample rate           &       100\,Hz &       200\,Hz \\
        Number of sensors     &           4 &           1 \\
        Steps per combination &          80 &          80 \\
    \midrule
        Total number of samples &       2880 &        8640 \\
        Walking duration      &      222\,min &           - \\ 
    \bottomrule
    \end{tabular}
    \label{tab1}
\end{table}

\section{Method}
\label{sec:Method}

Recurrent Neural Networks (RNNs) are a prevalent machine learning technique for dealing with time-series data. An RNN takes an external signal $\bm{x_t}$ and the previous hidden state $\bm{h_{t-1}}$ as input, and outputs the current hidden state $\bm{h_t}$ as:
\alignedequation{eq:RNN}{
    \bm{h}_{t} = f(\bm{W}\bm{x}_{t} + \bm{U}\bm{h}_{t-1})
}
where $\bm{W}$ and $\bm{U}$ are learnable parameters and $f$ is a non-linear activation function. One advantage of RNNs is their ability to deal with variable-length input sequences. RNNs are able to learn a distribution over a variable-length sequence by learning the distribution over the next input \cite{cho2014properties}. Early RNN models had difficulties dealing with long-term dependencies in data \cite{bengio1994learning, lipton2015critical}, which led to the development of architectures with more direct memory mechanisms including memory registers, and gated activation functions replacing simple non-linear activation functions.  In this paper we focus on two popular implementations; Long Short-Term Memory (LSTM) \cite{hochreiter1997long} and Gated Recurrent Unit (GRU) \cite{cho2014properties}.     

\subsection{Long Short-Term Memory}
\label{LSTM}
An LSTM cell (Fig.~\ref{fig:RNNs}a) comprises a memory cell $\bm{c_t}$, an input gate $\bm{i_t}$, a forget gate $\bm{f_t}$, and an output gate $\bm{o_t}$ as:
\begin{equation}
\begin{aligned}
  \bm{h}_{t} &= \bm{o}_t\tanh{(\bm{c}_t)} \\
  \bm{o}_{t} &= \sigma{(\bm{W}_o\bm{x}_t + \bm{U}_o\bm{h}_{t-1})} \\  
  \bm{c}_{t} &= \bm{f}_t\bm{c}_{t-1} + \bm{i}_t\bm{\tilde{c}}_{t-1} \\
  \bm{\tilde{c}}_{t} &= \tanh{(\bm{W}_c\bm{x}_t + \bm{U}_c\bm{h}_{t-1})} \\ 
  \bm{f}_{t} &= \sigma{(\bm{W}_f\bm{x}_t + \bm{U}_f\bm{h}_{t-1})} \\ 
  \bm{i}_{t} &= \sigma{(\bm{W}_i\bm{x}_t + \bm{U}_i\bm{h}_{t-1})}
\end{aligned}
\end{equation}

The input, forget, and output gates control how much new information is memorized, how much old information is forgotten, and how much information is output from the memory cell, respectively. 

\subsection{Gated Recurrent Unit}
\label{GRU}

\begin{figure}[b!]
\vspace{-0.5cm}
    \centering
    \includegraphics[width=8cm]{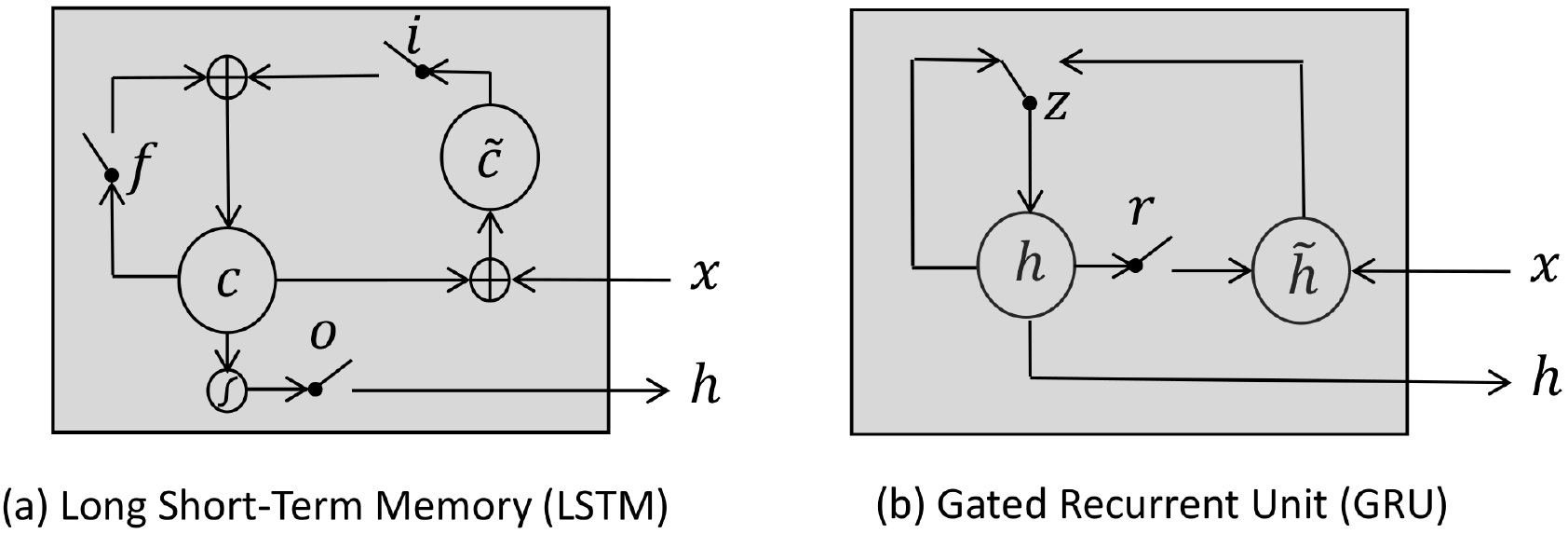}
    \caption{LSTM and GRU}
    \label{fig:RNNs}
\end{figure}

A GRU unit (Fig.\ref{fig:RNNs}b) consists of a reset gate $\bm{r_t}$, and an update gate $\bm{z_t}$, which reset and update the memory content adaptively. 
\begin{equation}
\begin{aligned}
  \bm{h}_{t} &= (1-\bm{z}_t)\bm{h}_{t-1} + \bm{z}_t\bm{\tilde{h}}_{t} \\
  \bm{z}_{t} &= \sigma{(\bm{W}_z\bm{x}_t + \bm{U}_z\bm{h}_{t-1})} \\  
  \bm{\tilde{h}}_{t} &= \tanh{(\bm{W}\bm{x}_t + \bm{U}(\bm{r}_t\odot\bm{h}_{t-1}))} \\
  \bm{r}_{t} &= \sigma{(\bm{W}_r\bm{x}_t + \bm{U}_r\bm{h}_{t-1})}
\end{aligned}
\end{equation}
where $\odot$ denotes element-wise multiplication. GRUs can be considered as an adaptive leaky integrator neuron \cite{gerstner2002spiking} in which $\bm{z_t}$ is a user-defined constant value. 

\subsection{RNN for Terrain Classification}
\label{Terrain Classification}

As shown in previous sections, GRUs have a simpler structure and fewer learnable parameters, and are therefore computationally more efficient compared to LSTM. However, it has been shown that RNN performance critically depends on the task and dataset \cite{chung2014empirical}. We evaluate both GRUs and LSTMs on multiple data sets in this paper. Prior work evaluated LSTMs on the PUT data set, \cite{bednarek2019touching},~\cite{bednarek2019robotic} but reports poor results on the raw data, producing predictions not much greater than random chance. Instead they developed methods in the frequency domain \cite{bednarek2019touching} with 80.39$\%$ accuracy, and fixed data windowing methods using a one dimensional Convolutional Neural Networks (CNNs) input layer \cite{bednarek2019robotic} fed into LSTMs which achieved 96.89$\%$, at the cost of data windowing. In our work we rigorously test both GRUs and LSTMs on the raw PUT dataset without windowing, training on multiple subsets of the data using $k$-fold cross-validation and reliably demonstrate high performance results using our RNN implementations provided as part of this paper. Prior work showed results for a single 90/10 train/validation split (without a holdout test data set or using cross-validation) in which the model may be overly biased and overfitted. We show comparable results using $k$-fold cross-validation (avoiding bias), without additional efforts to transfer or window the data. 



\begin{figure}[b!]
\vspace{-0.5cm}
    \centering
    \includegraphics[width=3in]{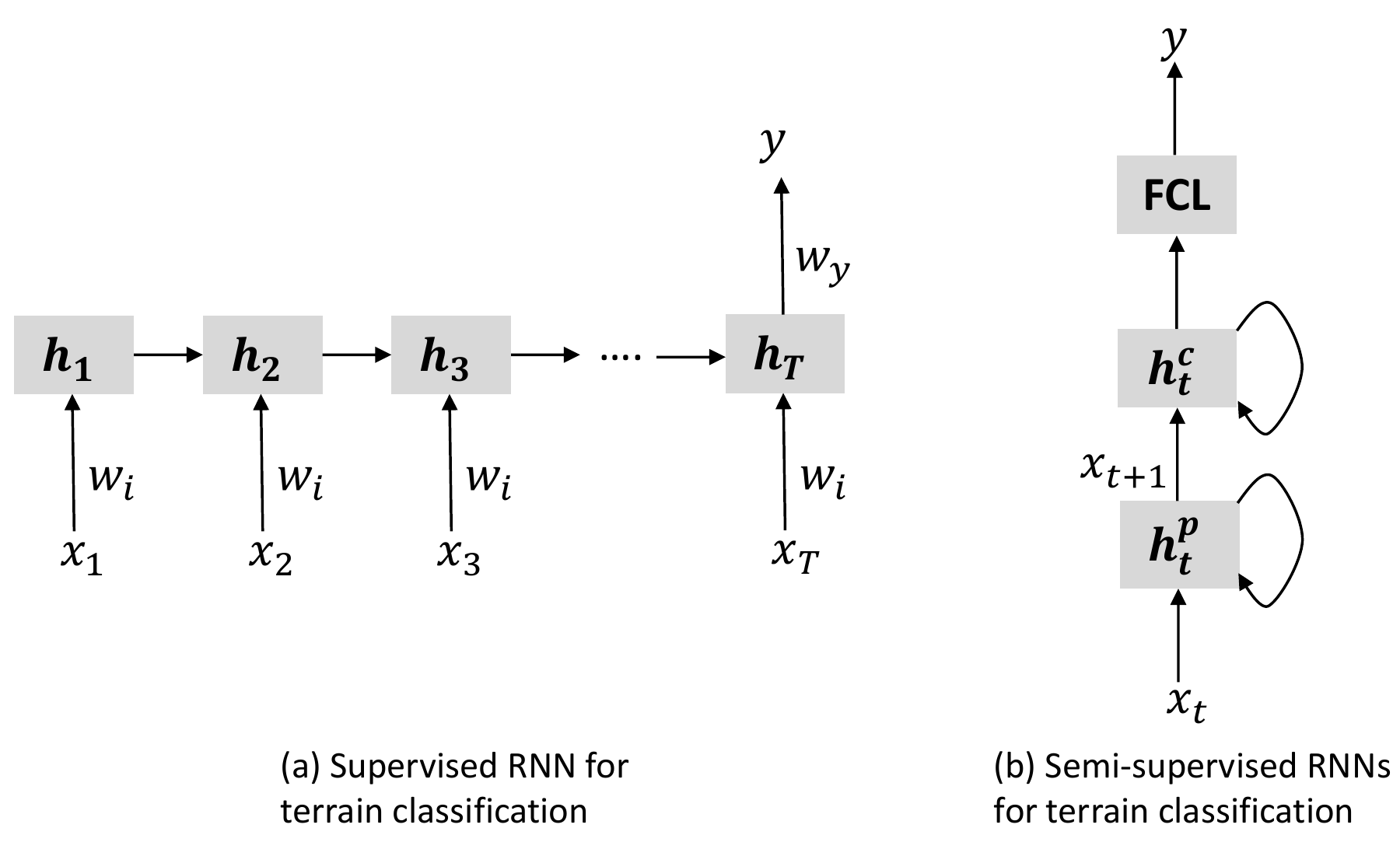}
    \caption{RNNs for terrain classification in a (a) Supervised fashion and (b) Semi-supervised fashion. Figure (a) shows an unfolded structure of the RNN model, while figure (b) illustrates the folded RNN models with cyclic representations. The $h_t^p$ and $h_t^c$ show the hidden states of the predictor RNN and the classifier RNN, which are stacked one above the other followed by a Fully Connected Layer (FCL).}
    \label{fig:Class}
\end{figure}

We also tested LSTMs and GRUs for the QCAT dataset to see whether gated RNNs can deal with a variable-length outdoor natural environment dataset. Fig.~\ref{fig:Class}(a) shows how an RNN can be used for this task. The model takes an external signal $\bm{x_{1:T}}$ from time step 1 to $\bm{T}$ during forward computation and only outputs $\bm{y}$ at time-step $\bm{T}$. The model does not need to output at every time-step because the whole $\bm{x_{1:T}}$ belongs to one class. This is different from a regression task where a regression model outputs a different signal at each time-step. As both datasets have variable-length sequences, $\bm{T}$ has a different value for each sequence. The output $\bm{y}$ is computed by softmax, which produces a distribution over the terrain classes. In addition, we tested semi-supervised RNNs for both datasets to examine their performances when only a small portion of annotated data was provided. The proposed model is shown in Fig.~\ref{fig:Class}(b), in which a classifier model is stacked above a predictor model. 


\section{Results}
\label{sec:Results}

We conducted our experiments to examine how gated RNNs perform in terrain classification tasks for legged robots in both supervised and semi-supervised fashions. The first experiment investigates GRUs and LSTMs for classifying terrains in the (indoor) PUT dataset, and the second experiment does the same on the outdoor QCAT dataset. The last experiment examines semi-supervised RNNs for both datasets. In all experiments, each dimension of the input data is normalized to zero mean and unit standard variation. The cross entropy between the ground truth data $\hat{\bm{y}}$ and the RNN output $\bm{y}$ was used for the loss function as:
\alignedequation{eq:Loss}{
    \bm{\bm{L}}_{loss} = \sum_{c=1}^{\bm{C}}{\hat{\bm{y}}_{c}\log{\bm{y}_{c}}}
}
where $\bm{C}$ is the total class number. An $\bm{L2}$ regularization term for $\bm{N}$ learnable parameters was also added to the loss as
\alignedequation{eq:Loss_Reg}{
    \bm{\bm{L}}_{loss} = \sum_{c=1}^{\bm{C}}{\hat{\bm{y}}_{c}\log{\bm{y}_{c}}} + \lambda \sum_{n=1}^{\bm{N}}{\bm{\theta}^2}
}

As neither data set has an independent set of test samples available, we apply $k$-fold cross-validation to reduce bias. Following commonly-used settings, we set $k$ to 5 and 10 in our experiments \cite{kuhn2013applied, james2013introduction}.

\subsection{RNNs Trained on the PUT Data Set}
We evaluated GRUs and LSTMs for classifying the terrains in the PUT dataset. The mean and Standard Deviation (SD) results of models trained by a simple loss function (Eq.~\ref{eq:Loss}) are shown in Table~\ref{tab2}. The Minimum (Min) accuracy among $k$ models and Maximum (Max) accuracy among $k$ models are also given. First, the most standard and straightforward architectures of GRUs and LSTMs are compared, then we added more complexity to the superior RNN to achieve higher accuracy. GRUs consistently outperform LSTMs, a trend which is repeated for 10-fold Cross-Validation (CV) models. Ten-fold models are noted to always have a better mean accuracy than 5-fold CV models. However, SD values are smaller for some of the 5-fold CV models. Table~\ref{tab3} illustrates the results of the models trained by loss with the $\bm{L2}$ regularization term (Eq.~\ref{eq:Loss_Reg}) in which $\lambda$ was set to 0.01. The performance of GRUs is improved by adding the regularization term, which is not always the case for LSTMs. One interesting performance is that even small GRUs (50 units) achieved a respectable 89.47$\%$ accuracy. This can be useful if one wants to use the classifier on embedded systems where memory efficiency is vital. The main result of note is that the more complex architecture (RNNs+FCLs) achieved a mean accuracy of $93.2\%$, which is significantly better (12.81$\%$) than the model reported in \cite{bednarek2019touching}, and $26.2\%$ better than the best results obtained by a conventional machine learning algorithm (SVMs), reported in \cite{bednarek2019touching}. The more complex architecture consists of GRUs plus Fully Connected Layers (FCLs) with dropouts~\cite{srivastava2014dropout}, and learning rate decay was used during its training phase. The notable network achieved $95.39\%$ maximum accuracy, whereas its minimum accuracy is $92.06\%$. This difference shows the importance of using cross-validation and reporting mean and standard deviation of all CV models. 

The overall results show that RNNs are capable of dealing with raw PUT data without reducing the temporal size or transferring from the time-domain to the frequency domain. Our results therefore suggest that gated RNNs are capable of dealing with long sequences specially when datasets are not too complex such as PUT. This also positions GRUs as a leading candidate for learning terrain classifiers in this context. We later show the importance of keeping the temporal resolutions for RNN models (Fig.~\ref{fig:PCA}).

Using a straightforward architecture, we show that our RNN implementation performs significantly better than previously reported results given the raw data\cite{bednarek2019touching, bednarek2019robotic}. For example, 50 LSTMs with no regularization and a fixed learning rate achieved a mean accuracy of $87.68\%$, compared to $18\%$ in the literature.

We suggest two possible reasons for poor performance observed in previous results. First, variable length sequences in the PUT dataset use zero-padding, which may perform poorly in certain (static) RNN implementations.  Secondly, the length of the sequence may not have been made available to the RNN, which forces common RNN implementations to revert from dynamic to static, causing the problem identified above.
 
 Specifically,  padded zeros affect both forward computation and back-propagation through time during training. In Eq.~\ref{eq:RNN}, the value of current hidden states $h_t$ depends on values of both input $x_t$ and previous hidden states $h_{t-1}$, therefore zero values of input $x_t$ will not necessarily result in zero values of current hidden states $h_t$. This means padded zeros can deteriorate the performance of RNNs if not handled properly.

\begin{table}[t!]
    \caption{ Accuracy for PUT dataset. SD, Min, Max, and CV abbreviate standard deviation, minimum accuracy among $k$ RNN models, maximum accuracy among $k$ RNN models, and cross validation, respectively.}
    \centering
    \begin{tabular}{cccccc}
    \toprule
           \multicolumn{2}{c}{}   & Mean\%   & SD\%  & Min\% & Max\%\\
    \midrule
           \multirow{6}{6em}{10-fold CV}  & 50 GRUs & 87.68  & 1.02 & 85.84 & 89.06\\
                      & 50 LSTMs & 83.68  & 1.06 & 81.35 & 84.99\\
                      & 100 GRUs & 90.15  & 1.11 & 88.30 & 91.42\\
                      & 100 LSTMs & 85.33  & 1.39 & 82.64 & 87.67\\
                      & 400 GRUs & 92.13 & 0.48 & 91.42 & 92.81\\
                      & 400 LSTMs & 90.43  & 0.59 & 89.6 & 91.43\\
    \midrule
           \multirow{6}{6em}{5-fold CV}  & 50 GRUs & 82.93  & 1.13 & 84.93 & 88.04\\
                      & 50 LSTMs & 81.86  & 0.76 & 80.47 & 82.73\\
                      & 100 GRUs & 90.2  & 0.6 & 89.49 & 90.94\\
                      & 100 LSTMs & 83.98  & 0.67 & 83.21 & 85.2\\
                      & 400 GRUs & 91.42  & 0.77 & 90.24 & 92.54\\
                      & 400 LSTMs & 89.67  & 0.74 & 88.95 & 91.1\\
    \bottomrule
    \end{tabular}
    \label{tab2}
\end{table}

\begin{table}[t!]
    \caption{ Accuracy for PUT dataset with regularization loss for training of networks.}
    \centering
    \begin{tabular}{cccccc}
    \toprule
           \multicolumn{2}{c}{}   & Mean\%   & SD\%  & Min\% & Max\%\\
    \midrule
           \multirow{6}{6em}{10-fold CV}  & 50 GRUs & 89.47  & 0.79 & 88.3 & 90.77\\
                      & 50 LSTMs & 83.63  & 1.85 & 81.22 & 87.34\\
                      & 100 GRUs & 90.71  & 0.87 & 89.7 & 92.49\\
                      & 100 LSTMs & 85.79  & 1.64 & 83.58 & 88.52\\
                      & 400 GRUs & 92.35  & 0.78 & 90.67 & 93.56\\
                      & 400 LSTMs & 88.3  & 1.49 & 85.74 & 91.0\\
                      & RNNs+FCL & \textbf{93.2} & 0.89 & 92.06 & 95.39\\
    \midrule
           \multirow{6}{6em}{5-fold CV}  & 50 GRUs & 88.44  & 1.06 & 86.86 & 90.08\\
                      & 50 LSTMs & 83.07  & 0.46 & 82.31 & 83.6\\
                      & 100 GRUs & 89.99  & 0.68 & 88.95 & 90.88\\
                      & 100 LSTMs & 85.44  & 0.38 & 85.09 & 86.11\\
                      & 400 GRUs & 91.91  & 0.41 & 91.26 & 92.38\\
                      & 400 LSTMs & 85.34  & 1.95 & 83.7 & 89.06\\
    \bottomrule
    \end{tabular}
    \label{tab3}
\end{table}

\begin{table}[t!]
    \caption{ Classification accuracy for QCAT dataset with regularization loss for training of networks.}
    \centering
    \begin{tabular}{cccccc}
    \toprule
           \multicolumn{2}{c}{}   & Mean\%   & SD\%  & Min\% & Max\%\\
    \midrule
           \multirow{6}{6em}{10-fold CV}  & 50 GRUs & 87.71  & 2.28 & 84.37 & 92.01\\
                      & 50 LSTMs & 80.56  & 0.96 & 78.82 & 81.94\\
                      & 100 GRUs & 89.58  & 1.24 & 88.19 & 92.36\\
                      & 100 LSTMs & 82.29  & 1.15 & 80.56 & 84.38\\
                      & 350 GRUs & 93.02  & 1.62 & 90.97 & 96.53\\
                      & 350 LSTMs & 85.97  & 1.64 & 82.29 & 87.85\\
                      & RNNs+FCL & \textbf{96.6} & 0.89 & 95.49 & 98.61\\
    \midrule
           \multirow{6}{6em}{5-fold CV}  & 50 GRUs & 86.6  & 1.54 & 85.07 & 89.41\\
                      & 50 LSTMs & 81.46  & 1.64 & 79.17 & 83.33\\
                      & 100 GRUs & 87.71  & 1.03 & 86.46 & 88.72\\
                      & 100 LSTMs & 83.82  & 1.67 & 80.9 & 85.42\\
                      & 350 GRUs & 91.28  & 1.05 & 90.1 & 92.88\\
                      & 350 LSTMs & 83.19  & 1.21 & 81.25 & 85.07\\
    \bottomrule
    \end{tabular}
    \label{tab5}
\end{table}

\subsection{RNNs Trained on the QCAT Data Set}

We evaluated GRUs and LSTMs for classifying the terrains on the outdoor, variable-length QCAT dataset. Models were trained on data collected by the force sensors (OptoForce sensors), and the IMU (XSens). IMUs measure  movement and orientation, which is affected by many factors other than the surface it walks on. It is also susceptible to noise, which may make the data harder to work with. Force sensors on the feet are more expensive and mounting them can be mechanically and electrically challenging. They are not available on many platforms due to this, but the advantage is that they are in direct contact with the surface, which may reduce noise. 

Table~\ref{tab5} illustrate the results achieved by GRUs and LSTMs using four force sensors and trained via regularized loss (Eq.~\ref{eq:Loss_Reg}).  Overall, results follow similar trends to those observed with the PUT dataset. GRUs again outperforms LSTMs, and the best network (96.6$\%$ of accuracy) belongs to 10-fold CV models with FCL.

\begin{table}[b!]
    \caption{ Accuracy for QCAT dataset with IMU sensor with regularization loss for training of networks..}
    \centering
    \begin{tabular}{cccccc}
    \toprule
           \multicolumn{2}{c}{}   & Mean\%   & SD\%  & Min\% & Max\%\\
    \midrule
           \multirow{6}{6em}{10-fold CV}  & 50 GRUs & 90.07  & 1.77 & 87.5 & 93.06\\
                      & 50 LSTMs & 83.06  & 2.46 & 78.82 & 87.5\\
                      & 100 GRUs & 93.06  & 1.27 & 91.32 & 95.49\\
                      & 100 LSTMs & 84.44  & 2.52 & 78.47 & 88.19\\
                      & 350 GRUs & 94.37 & 1.68 & 90.97 & 96.88\\
                      & 350 LSTMs & 85.0  & 2.11 & 81.25 & 88.19\\
                       & RNNs+FCL & \textbf{96.63} & 1.17 & 94.44 & 98.26\\
    \midrule
           \multirow{6}{6em}{5-fold CV}  & 50 GRUs & 88.61  & 1.11 & 87.15 & 90.28\\
                      & 50 LSTMs & 80.83  & 1.43 & 78.47 & 82.64\\
                      & 100 GRUs & 90.63  & 1.29 & 88.19 & 92.01\\
                      & 100 LSTMs & 83.23  & 1.02 & 81.6 & 84.55\\
                      & 350 GRUs & 92.7  & 0.87 & 91.32 & 93.92\\
                      & 350 LSTMs & 84.03  & 2.17 & 82.12 & 87.33\\
    \bottomrule
    \end{tabular}
    \label{tab7}
\end{table}

We also consider a robot that relies either on a single force sensor, or solely on the IMU. As discussed earlier, force sensors are expensive and many platforms do not have them, demonstrating our method working with a more accessible, reduced sensory payload therefore has merit. 

We achieved a mean accuracy of $87.74\%$ when using only one force sensor. Using a cheap IMU sensor (Table~\ref{tab7}), we see remarkably high mean accuracy ($96.63\%$) that is slightly better than the one achieved by the four force sensors ($96.6\%$). The results suggest that GRUs are capable of dealing with noise in the input signal, showing that our classification models are widely applicable across many robot platforms. Trends are consistent with previous results obtained using force sensors. Notably, more of the 5-fold CV models had less variance than the 10-fold CV models. We also investigated the models given both IMU and force inputs and observed that they could achieve the accuracy of $97.64\pm{1.03}\%$. Given the remarkable accuracy obtained by IMU or force sensors, it does not seem necessary to have both types of sensors for robotic terrain classification, although more investigations are required to ascertain the limits of this, or applicability to different types of robots.     

\subsection{Semi-supervised learning}
As shown for both PUT and QCAT datasets, supervised RNNs achieved high accuracy for classifying the different terrains. However, the process of data collection for supervised models is tedious. For data annotation, either the robots need to walk on different types of terrains separately or one needs to hand-label the data later on. The effort can be reduced if only a portion of data needs to be annotated for comparable performance. To that end, we proposed semi-supervised RNNs for the terrain classification to investigate the possibility of reducing hand-labeling efforts.

\begin{figure}[t!]
    \centering
    \subfigure[PUT results]{\includegraphics[width=75mm]{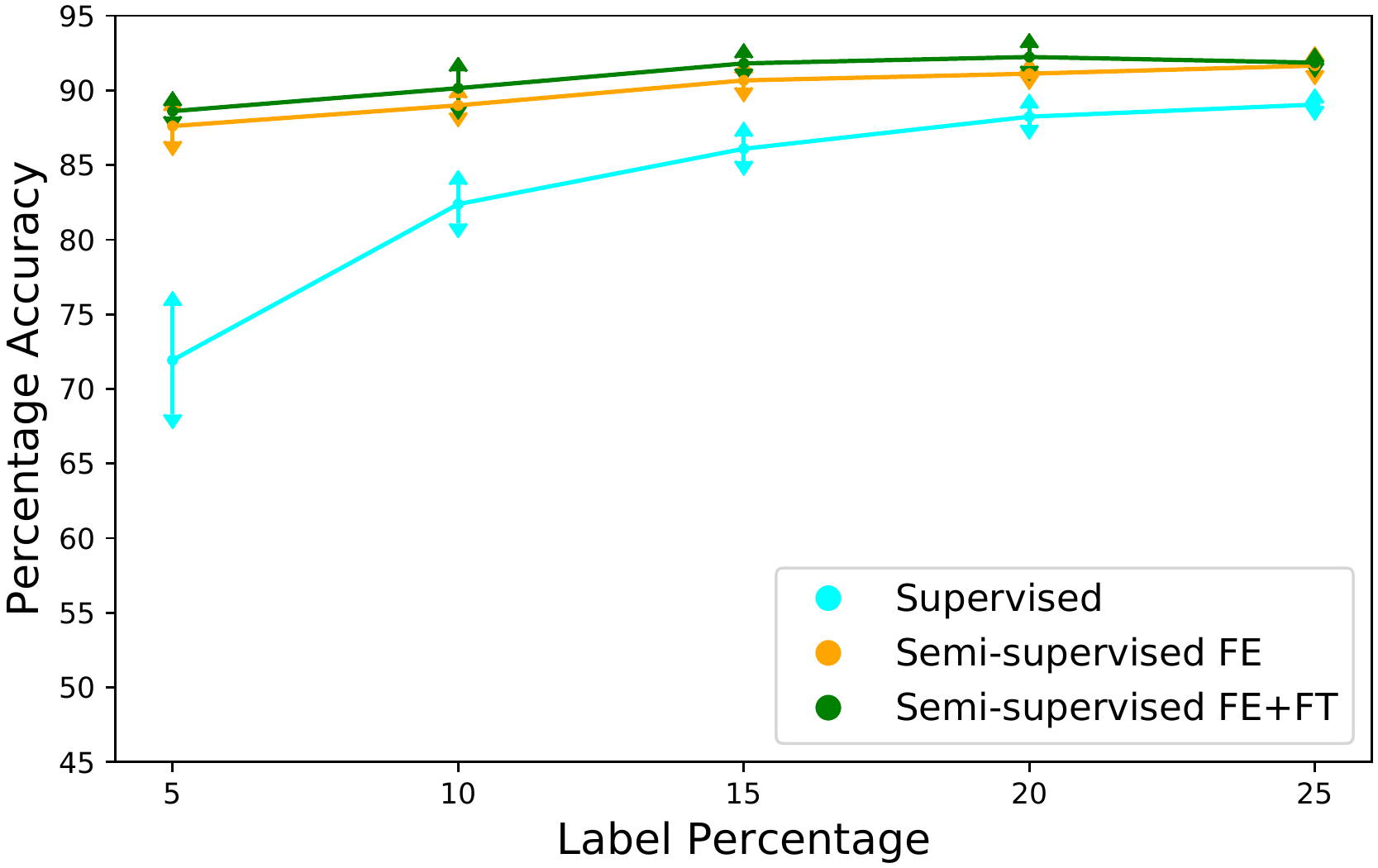}}
    \subfigure[QCAT results]{\includegraphics[width=75mm]{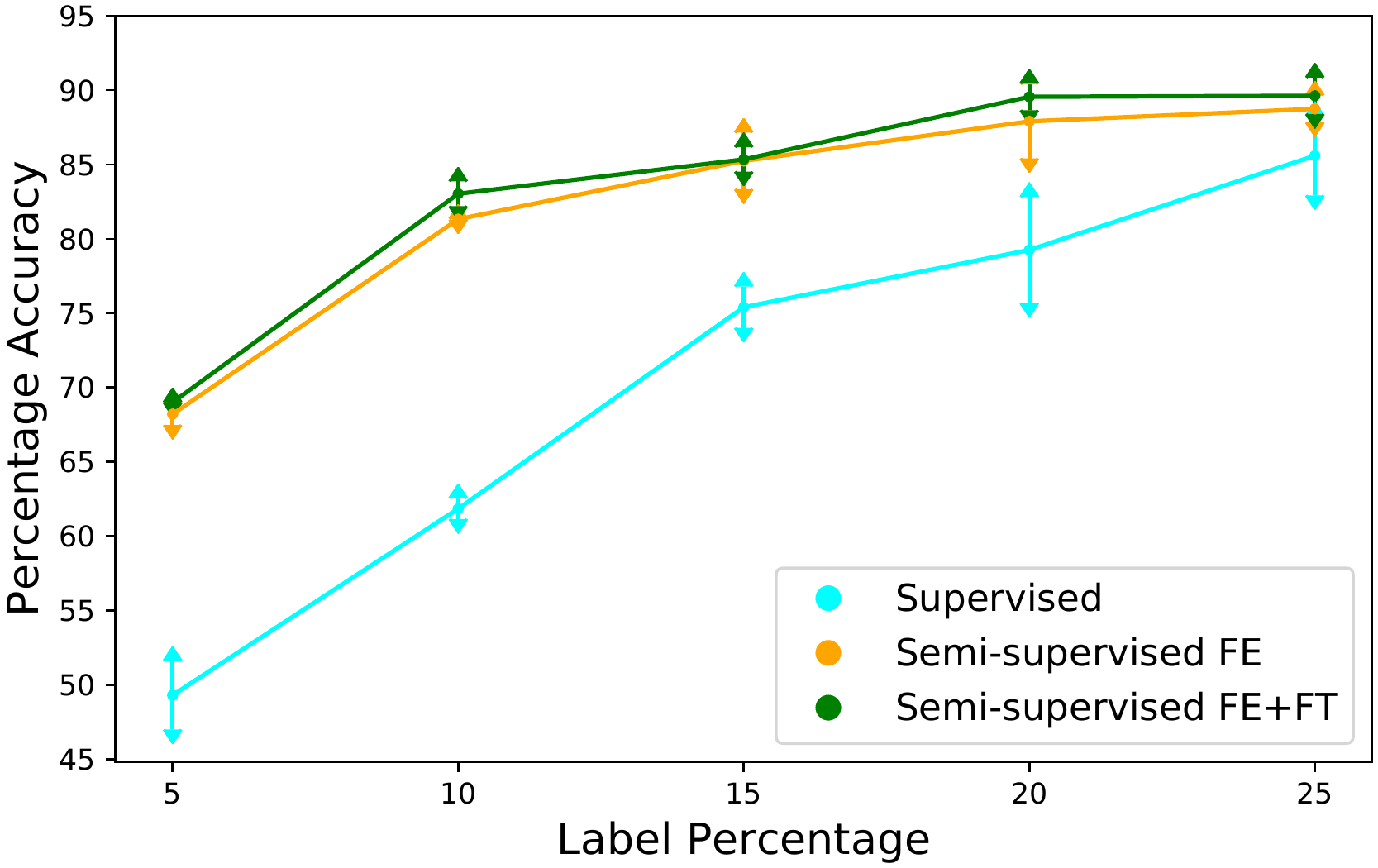}}
    \caption{Comparison of test results between supervised learning and two types of semi-supervised learning. The arrows show the standard deviations. The semi-supervised algorithms significantly outperform the supervised algorithm for both datasets, specially when the percentage of the label data is small. The semi-supervised learning with feature extraction and fine tuning has a superior performance than the semi-supervised one with only feature extraction.}
    \label{fig:semi_comp}
\end{figure}

In the proposed semi-supervised model, we stacked unsupervised RNNs, supervised RNNs, and Fully Connected Layer (FCL) neural networks (Fig.~\ref{fig:Class}(b)). First, the unsupervised RNNs, also referred as the predictor RNNs, are trained by taking the current signal $\bm{x_t}$ (Force/Torque signals for PUT and IMU signals for QCAT) as input and predicting the next step signal $\bm{x_{t+1}}$ in their outputs. A portion of input data (predicting data) was used for training of the predictor RNNs. This method of training is called unsupervised learning because the target signals are provided by input signals themselves and no human efforts for data annotations are needed. After training, the weights are frozen and the supervised RNNs, also referred as the classifier RNNs, and the FCL are trained by using other portion of the data (classifying data). The classifier models take the predictor RNN outputs as input and estimating the outputs $\bm{y}$, types of terrains, at time-step $\bm{T}$. We refer to this model as Feature Extracting (FE) semi-supervised learning because the predictor RNNs were used as feature extractor. We also investigated fine-tuning of the whole FE semi-supervised learning model by decreasing the learning rate and retraining the whole model with classifying data. This model is referred as Feature Extracting (FE) + Fine Tuning (FT) semi-supervised learning model. Both models are compared on PUT and QCAT datasets. For sake of space and given that GRUs outperformed LSTMs in all previous experiments, only GRUs are used in the following experiments. IMU data is only used for the QCAT dataset given our previous promising results with this cheap, available sensor.  

We investigated 5 different splits of predicting/classifying data. In Fig.~\ref{fig:semi_comp}, $5\%$ means that the whole dataset was randomly divided into two sections: $90\%$ of data was used as the predicting data and the remainder was  classifying data. Using 2-fold cross-validation to lower classifier bias, the classifying data ($10\%$ of the whole data) was randomly split into 2 parts. Therefore, the supervised model (classifier RNNs and FCL) was only trained by $5\%$ of the whole data, and evaluated on the other $5\%$. As $5\%$ of the data was used for training of the supervised model, we called these models '$5\%$'. We tested the classifiers on the rest $90\%$ of the whole data, which are referred as test accuracy. Similarly, '$25\%$' means that the whole dataset was randomly divided into two sections: $50\%$ predicting data and $50\%$  classifying data. All networks were trained by $25\%$ of the data, evaluated on $25\%$, and tested on $50\%$. In order to compare the semi-supervised models with supervised models and show the effect of the predictor RNNs in the semi-supervised models, we trained supervised classifiers using the classifying data.

\begin{figure*}
    \centering
    \includegraphics[width=150mm]{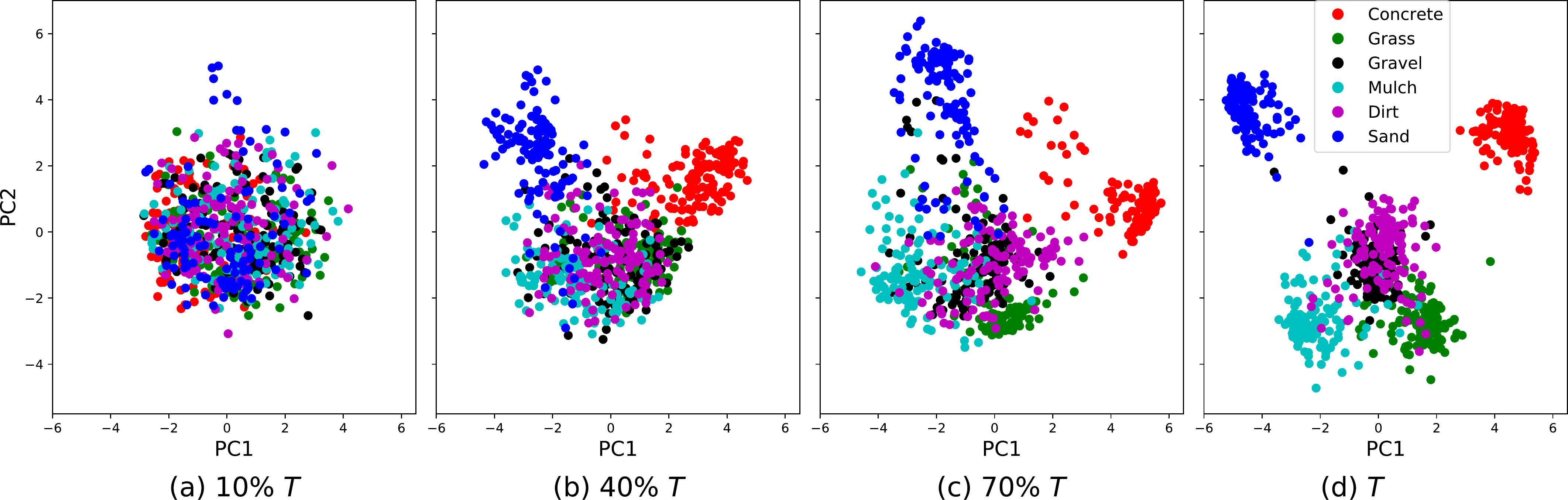}
    \caption{Principle component analysis of GRU hidden states given 720 testing samples at different time-steps. The points belonging to the same class have clustered together as time evolves. Some classes (sand and concrete) seem to be easier for the RNN model to classify than others, specifically mulch, dirt, and gravel.}
    \label{fig:PCA}
\end{figure*}

Test accuracy results (Fig.~\ref{fig:semi_comp}) shows the superior performance of semi-supervised models over supervised models. The gap between their percentage accuracy is significantly larger for a smaller amount of labels, and becomes smaller as more labels are provided during training. Supervised model accuracy are relatively low ($49.31\%$ and $61.85\%$) when networks are trained by $5\%$ and $10\%$ labels, possibly because QCAT is a relatively small dataset. The effect is less evident for PUT, which is larger. 

Results show the effectiveness of fine-tuning for semi-supervised learning, as those models outperform the FE semi-supervised models in all cases. Accuracy is slightly lower for $25\%$ ($92.24\%$ accuracy) than $20\%$ ($91.86\%$ accuracy) for PUT dataset, and it is really close for QCAT. This is likely because predictor RNNs are trained on less data ($50\%$) for the former model, although its classifier model had access to $5\%$ more labels, suggesting that a balance between amounts of training data for the supervised and unsupervised model may be needed to achieve the highest possible accuracy using semi-supervised models. The overall results indicate that semi-supervised learning is an effective learning method when few labels are available, addressing a major shortcoming of supervised learning, especially for terrain classification where extensive annotations may be difficult to procure .

We visualised the temporal evolution of hidden states of a GRU given 720 testing samples of different lengths ($\bm{T}$). Principal Component Analysis (PCA) reduced the GRU hidden state dimension (200 GRU units) to two principle components. Fig.~\ref{fig:PCA} displays the PCA results at time-step $\bm{t}$ = $10\% \bm{T}$, $40\% \bm{T}$, $70\% \bm{T}$ and $\bm{T}$ for QCAT dataset. The dirt, mulch, and gravel classes are seen to bethe most challenging ones for the network due to similarities between these three classes (Fig.~\ref{fig.terrains}). Sand and concrete are the most separable classes for the network due to their contrasting mechanical properties. The figure also shows how the hidden states of the same classes are mostly clustered together as time goes by. This emphasizes the importance of the data temporal resolution for the terrain classification task. 

We investigated other possible architectures for semi-supervised learning, which have been shown to be effective in tasks such as image classification or NLP. Our first approach replaced predictor RNNs with fully connected layers. Second, we removed the predictor RNNs, pre-trained RNNs and fully connected layers for the one-step prediction task, and fine-tune the whole network for the classification task. We also replaced the predictor RNNs with auto-encoder RNNs, meaning that the unsupervised RNNs were trained on the current signal $\bm{x_t}$ (Force/Torque signals for PUT and IMU signals for QCAT) and predicted the the same time step signal $\bm{x_{t}}$.  Preliminary results suggest that non of the aforementioned semi-supervised methods are effective for the robotic terrain classifications (PUT and QCAT), they under-performed the supervised models, although more detailed analysis and investigations are needed to have a conclusive statement.


\section{Conclusions}
\label{sec:conclusions}
We utilized supervised and semi-supervised gated RNNs for the robotic terrain classification. Our classifiers were first evaluated on the PUT  dataset composed of time-series data with variable lengths that were collected in an indoor environment. RNN models given time-series dataset significantly exceeded the accuracy rates of the SVM and the fully connected neural model using frequency-domain transferred data. Furthermore, we achieved high accuracy rates by RNN classifiers on our own dataset composed of time-series data with variable lengths that were collected in an outdoor environment. The results obtained from both datasets suggest that a GRU outperforms LSTM for proprioceptive terrain classification. The results also show the importance of the data temporal resolution for terrain classification. In the second experiment, we showed that IMU sensors, available on many robot platforms, may be a sensor of choice for terrain classification. We introduced the first deep semi-supervised models for robotic terrain classification, and showed that they are capable of directly dealing with raw and variable-length time-series data.  Results indicate that semi-supervised models outperformed supervised models remarkably when only small amounts of annotated data are available, suggesting that less annotated data is required for terrain classification, and thus larger usable data sets can be easily made available. The results open up an interesting future extension of the current work: a transfer learning between robot platforms, meaning that a different robot will be used for training the unsupervised model with no annotated data. Another interesting future work can be to investigate probabilistic RNNs as semi-supervised models for robotic terrain classification. Probabilistic RNNs \cite{bayer2014learning, chung2015recurrent, ahmadi2019novel} are known to attain better generalization capabilities than deterministic ones specially for dealing with data with high variability as evidenced in this domain. 



\section*{Acknowledgements}
We  would  like  to  give  our  special  thanks  to  people  who  helped  us  with  the  current study. we are particularly grateful for great assistant given by Benjamin Tam for data collection. 

\balance

\bibliographystyle{IEEEtran}
\bibliography{main.bbl}

\end{document}